# A Data Driven Method of Optimizing Feedforward Compensator for Autonomous Vehicle Control

Tianyu Shi[1], Pin Wang[2*], Chonghao Zou[1], Long Xin[3*], Ching-Yao Chan[2]

*Abstract*—A reliable controller is critical for execution of safe and smooth maneuvers of an autonomous vehicle. The controller must be robust to external disturbances, such as road surface, weather, wind conditions, and so on. It also needs to deal with internal variations of vehicle sub-systems, including powertrain inefficiency, measurement errors, time delay, etc. These factors introduce issues in controller performance. In this paper, a feed-forward compensator is designed via a data-driven method to model and optimize the controller's performance. Principal Component Analysis (PCA) is applied for extracting influential features, after which a Time Delay Neural Network is adopted to predict control errors over a future time horizon. Based on the predicted error, a feedforward compensator is then designed to improve control performance. Simulation results in different scenarios show that, with the help of with the proposed feedforward compensator, the maximum path tracking error and the steering wheel angle oscillation are improved by 44.4% and 26.7%, respectively.

## I. INTRODUCTION

Recently, Autonomous Vehicles (AVs) have attracted much attention with potential improvement in driving safety, transportation efficiency and fuel economy. A typical autonomous system features several main functional layers: perception, decision-making, path planning and control. The control layer guarantees the vehicle performance of tracking the desired command inputs, e.g. velocity and yaw rate pairs $(v_d, w_d)$. The measured velocity and yaw rate pairs $(v, w)$ affect the upper layers (e.g. decision and planning layers) and in turn influence the control command inputs [1]. As bucket effect reveals, even if we have proper functions in upper layers, the overall performance of the autonomous vehicle will not be optimal without solid low-level control layers. Therefore, learning and modeling the vehicle low level controller's performance based on real-world driving data is vitally important in the development of an autonomous vehicle system. Currently, most autonomous driving experiments adopt the autonomous driving platforms with a drive-by-wire system by controlling throttle paddle $T_t$, brake paddle $T_b$ and steering wheel angle δ to track the desirable waypoints of a given trajectory.

The main challenge in controller design lies in the difficulties of modeling the vehicle systems with uncertainties. One challenge is that the access to engine, brake, and steering systems is through several electronic control units in low-level control, which is unrevealed to the drive-by-wire system. The other challenge is that it is hard to obtain a precise dynamic model due to the inherent coupling of dynamics of multiple sub-systems and highly non-linearity of the vehicle [2, 3, 4].

Therefore, learning and modeling the vehicle low-level controller's performance based on real-world driving scenarios is vitally important for development of an autonomous vehicle system.

Data driven methods have a great potential for nonlinear system prediction. For instance, Neural Network (NN) has an excellent capacity of mapping complex inputs and outputs by training with a large amount of off-line data. Some researchers proposed adaptive PID control and used neural network to optimize the existent controller's parameter in autonomous vehicle [5, 6]. However, sometimes the application is limited, because the labeling signal of supervised learning is hard to obtain due to environmental noise. In addition, the optimized result is unstable, and inappropriate for implementation in the low-level controller. Some researchers used neural network in vehicle dynamic modeling. Wang *et al.* [7] proposed to obtain the data by measuring the dynamic parameters of vehicle, then set the mathematical model to predict the engine torque and vehicle speed using non-linear neural network. However, they did not make it explicit how and why they chose these data for the input nor further validation of whether their method could optimize the controller's performance. Neural networks have been wildly applied to enhance autonomous driving performance [8, 13, 14]. Wang *et al.* [8, 15] proposed a lane change control model under continuous action space based on reinforcement learning algorithm, but how the output action was converted to a low-level controller was not addressed.

Unlike previous research, we mainly focus on developing an error mapping model between the designated input commands and actually measured outputs of vehicle dynamics, based on a data-driven method. By building the input-output model, we predicate errors between input and output, which can be applied to the feedforward compensation optimization.

Our main contribution in this paper is that we designed a feedforward compensator based on the prediction for future control errors. It is considered to be more reliable and safer for optimizing the motion control of autonomous vehicles as there is no need to change the basic structure of the drive-by-wire controller.

 The rest of the paper is organized as follows: Section II presents the methodology of how Principle Components Analysis (PCA) and Time Delay Neural Network (TDNN) is used in our study. Section III introduces the data that we used in developing the error compensation model. Section IV illustrates the results in detail. Section V summarizes the major contributions and concludes the paper.

## II. METHODOLOGY

Our goal is to derive a mathematical model of a system using observed data. The architecture of the methodology is

T. Shi and C. Zou are with Beijing Institute of Technology, Haidian, Beijing, 100081, China. (emails: tianyu.s@outlook.com, zouchonghaobit@163.com)

P. Wang and C. Chan are with California PATH, University of California, Berkeley, Richmond, CA, 94804, USA. (emails: {pin_wang, cychan} @berkeley.edu)

L. Xin is with School of Vehicle and Mobility, State Key Lab of Automotive Safety and Energy, Tsinghua University, Haidian, Beijing, 100084, China. (email: xin-l13@mails.tsinghua.edu.cn)

*P. Wang and L. Xin are corresponding authors.

depicted in Fig. 1. The set of influential features are defined based on the prior knowledge of the data collection experiment. Principal features are obtained through Principal Component Analysis (PCA). Error compensation model and the feedforward model are then built based on these features.

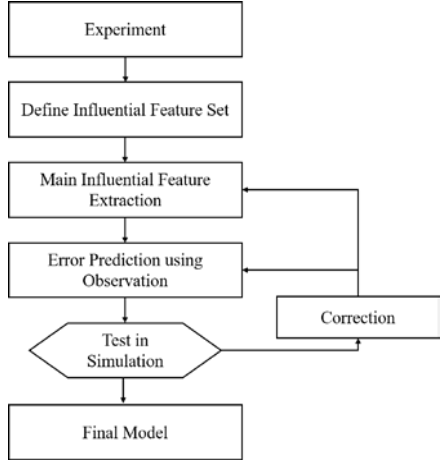

Figure 1. Architecture of the propose method

*A. Key Factors Extraction*

The control errors are sensitive to various variables, such as steering wheel angle, vehicle speed, speed feedback of wheels, etc., which leads to a complicated situation and influence the design of an error compensation model.

In our training data, the number of samples collected from on-board sensors is $m$, and the number of features is $n$. The input features include velocity, angular velocity and acceleration for all three axes, as well as steering wheel angle, torque, and speed for all four wheels. The data matrix $X$ is established as follows:

$$X = \begin{bmatrix} x_{11} & \cdots & x_{1n} \\ \vdots & \ddots & \vdots \\ x_{m1} & \cdots & x_{mn} \end{bmatrix} \quad (1)$$

where each row represents a set of experimentally obtained data. In order to apply PCA, we perform feature centering by calculating the mean and covariance matrix of the sample of each dimension.

The principal component is determined by calculating the contribution rate of different components. Based on principal component analysis, the contribution rates are sorted from high to low, and the principal components corresponding to the top 3 eigenvalues that satisfy our required contribution rate are selected. The formula of the contribution rate $Cr$ is as follows:

$$Cr = \frac{\lambda_i}{\sum_{i=1}^{n} \lambda_i} \quad (2)$$

The eigenvectors $v = (v_1, v_2, \ldots, v_n)$ corresponding to the eigenvalues $\lambda = [\lambda_1, \lambda_2, \ldots \lambda_n]^T$. Finally, we select the steering wheel angle, steering wheel torque and longitudinal velocity as our main input features to the network. (Detailed computation result can be seen in *data analysis* of section III)

*B. Error Estimation*

The control system for autonomous vehicle is a highly complex hysteresis nonlinear dynamic system. We will elaborate on this aspect further in Section III. By considering the correlation of previous and current features, we adopt Time Delay Neural Network (TDNN) as our training model which can ensure that the output of our network has previous information. As a result, we can model the hysteresis characteristic of the dynamic system.

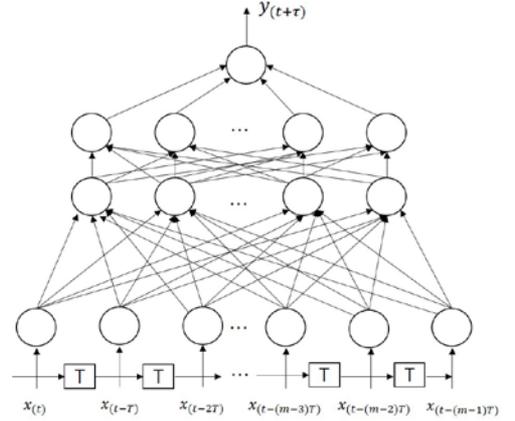

Figure 2. Time Delay Network structure

The prediction model can be established in the following process. The structure of the neural network is represented in Fig. 2. The input vectors $[x(t), x(t-T), \ldots, x(t-(m-1)T)]$ are the current and previous features of main influential factors, and each vector includes steering wheel angle, steering wheel torque and longitudinal velocity, which are selected by PCA. The output $y(t+\tau)$ represents the predicted error $\hat{e}(t+\tau)$ between the designated command input and the actual output in the future. The network has two hidden layers with 8 nodes and 6 nodes in each layer. For the activation function, we use a *tansig* function as follows:

$$y_j = tansig(x_j') = \frac{2}{1+e^{-2x_j'}} - 1 \quad (3)$$

We use the square error function as our loss function:

$$e(w, b) = \frac{1}{2n} \sum_{i=1}^{n} \left( y_{mea}(x_i) - y_{(t+\tau)}(x_i) \right)^2 \quad (4)$$

where $e(w, b)$ is the loss function, $x_i$ is the i-th sample, n is the total number of samples, $y_{mea}(x_i)$ is the measured steering wheel angle error between the input and output, while $y_{(t+\tau)}(x_i)$ is the predicted error based on neural network.

The learning rate is 0.001 and the final output is the predicted error of the steering wheel angle between command input and measured output.

*C. Feedforward Compensator Design*

The compensate process is depicted in Fig. 3. The reference trajectory for path tracking is given as the input to the system. Then, the path tracking control (PTC) approach will generate several time series $u_2(t)$ as command input to the controlling plant (P). From the errors between the command input $u(t)$ and the actual output $\theta(t)$ as well as the correlation between current and previous data information acquired from on-board sensors, the TDNN network can learn to forecast the future compensation error $\hat{e}(t+\tau)$ for the next $(t+\tau)$ time steps.

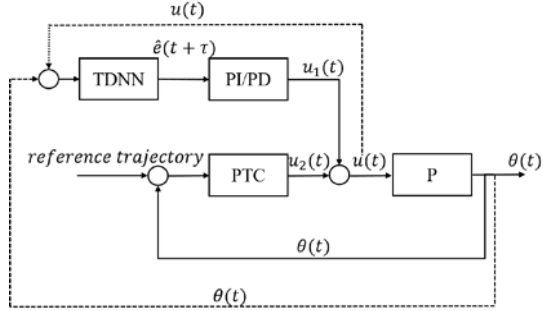

Figure 3. Compensator architecture

To optimize the control performance of the autonomous vehicle, we have designed a PI controller and a PD controller for different situations. The relationship between $\hat{e}(t+\tau)$ and $u_1(t)$ is illustrated in (5). In the equation, $T$ is the sampling period ($T = 0.05$). $w_0$ is the threshold ($w_0 = 2\ deg/s$). Output $u_2(t)$ is added into the command input $u_1(t)$ to generate the compensated input $u(t)$.

$$\begin{cases} u_1(t) = (k_p + k_i \dfrac{Tz}{z-1})\hat{e}(t+\tau), & |\gamma| < w_0 \\ u_1(t) = (k_p + k_d \dfrac{z-1}{Tz})\hat{e}(t+\tau), & |\gamma| > w_0 \end{cases} \quad (5)$$

When the desired yaw rate γ is within the threshold, we can assume the vehicle drive on a straight road and the predicted error $\hat{e}(t+\tau)$ goes through the PI controller to generate the processed signal $u_1(t)$. The output of the PI controller will be reset to zero when the predicted error $\hat{e}(t+\tau)$ crosses zero. This can make the steering control on a straight road more stable and mitigate steering angle oscillation when the vehicle proceeds on a straight road. When the yaw rate is beyond the threshold, we can assume the vehicle drives on a curved road. The predicted error $\hat{e}(t+\tau)$ will go through the PD controller, generating the processed signal $u_1(t)$ to compensate the $u_2(t)$ to optimize the control stability of the plant. With the help of the PD controller on a curved road, the compensator can generate quick response and enhance the control performance.

## III. EXPERIMENT DATA

### A. Data Collection

We use Lincoln MKZ, equipped with different sensors and software, to collect real-world driving data, as shown in Fig 4.

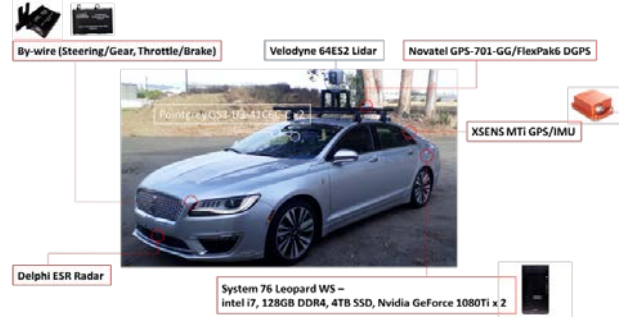

Figure 4. Senosors and computer inplemented on testing platform

In the vehicle platform, the drive-by-wire system receives the sensor data over the CAN bus and transmits the data to the planning layer. The planning layer will generate the desired commands to actuators of the vehicle. The steering control is implemented with a feedforward proportional controller, and the yaw rate and current speed measurement are used to compute a nominal steering angle based on a kinematic bicycle model. The steering control work flow is shown in Fig 5.

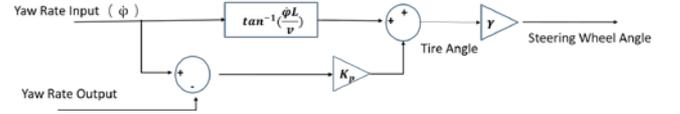

Figure 5. Steering control work flow

The test was conducted in Richmond Field Station, UC Berkeley. The driving scenarios include straight road, U-turn, and road segment with slopes. These represent typical road types in real-world driving conditions in suburban areas. With the use of the preview target waypoints for path tracking, the upper layer can generate desired input velocity and yaw rate pairs $(v_d, w_d)$ into the lower level controller.

### B. Data Analysis

The experimental vehicle is built on the robot operation system (ROS) [10], with which we can collect data of the vehicle controller performance. We basically record the data from the IMU, GPS and CAN bus. The data elements include orientation of x, y, z, w, linear acceleration of x, y, z, steering wheel angle command, steering wheel angle output, steering wheel torque, etc. After preprocessing the data collected from sensors, we analyze the performance of the control system.

First, we calculated the contribution rate based on PCA as shown in Table I. We can select 'steering wheel angle, steering wheel torque, longitudinal velocity' as our main input features.

TABLE I
THE CALCULATED COMPUTATION RATE OF FEATURES

| Type of data read by Sensors | Eigenvalues | Cr |
|---|---|---|
| Steering wheel angle | 117.383 | 47.80% |
| Velocity( x axis) | 116.385 | 47.39% |
| Steering wheel torque | 11.674 | 4.75% |
| Turning radius | 0.033 | 0.01% |
| Linear acceleration (x,y,z axis) | 0.032 | 0.01% |
| Veloctiy ( y,z axis) | 0.03 | 0.01% |
| Angular velocity (x,y,z) | 0.027 | 0.01% |
| Front right wheel speed | 0.021 | 0.01% |
| Front left wheel speed | 0.004 | 0.00% |
| rear right wheel speed | 0.004 | 0.00% |
| Rear left wheel speed | 0.001 | 0.00% |

As is shown in Fig. 6, during the drive on a straight road (steering wheel angle between -0.2rad~0.2rad), the $RMSE_{straight} = 0.0581$. As for the curved road (absolute value of steering wheel angle more than 0.2rad) the $RMSE_{curve} = 0.3218$. We can conclude that the error is relative larger in the curved route. We tested the curve driving in simulation. We simulated 'double lane change' scenario [11] and analyzed the steering stability of driving on a curved road.

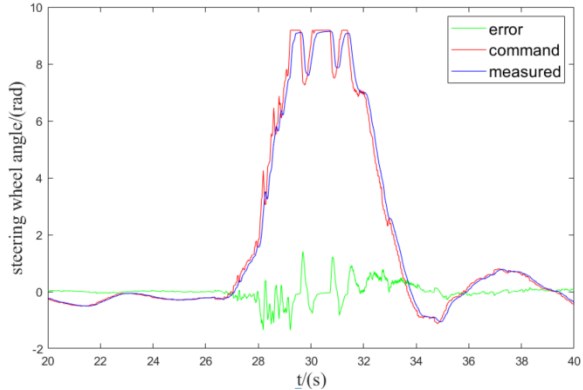

Figure 6.  Steering control performance (selected from 20s to 40s)

We find that there are errors between the expected output in the control system and the measurement output of the system. Take the steering wheel angle as an example, as shown in Fig. 6, the error (here we use RSME to evaluate) between the steering angle during the whole test is 0.254. It can be seen that the output response always falls behind the command input because of the time delay τ. To quantify the response time delay τ, we translate the command input Δt ( Δt =0.02s, 0.04s,0.06s, …, 0.40s) to the right, then measure the RMSE between the command input and measured output. The result is shown in Fig 7. From the trend of RMSE in the Fig 7, we can conclude that in Δt =0.2s, the smallest RMSE=0.0713 (28.7% of the RMSE when Δt =0s) which indicates that the error between the command input and the measured output of the steering wheel angle is the minimum. According to the above analysis, we can quantify the delay time as 0.2s. After compensating the delay time, we find that 71.3% of the error is due to the steering control time delay.

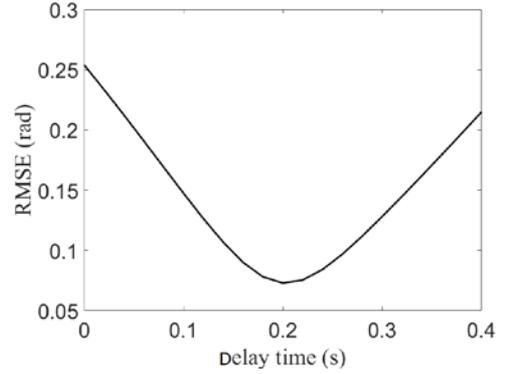

Figure 7.  Time delay analysis

## IV. DEVELOPED MODELS

### A. Error Estimation Model

The training process with the neural networks is conducted in matlabR2018b. In the error estimation process, we tested four typical networks, BP network, TDNN network, NARX network, and LSTM network.

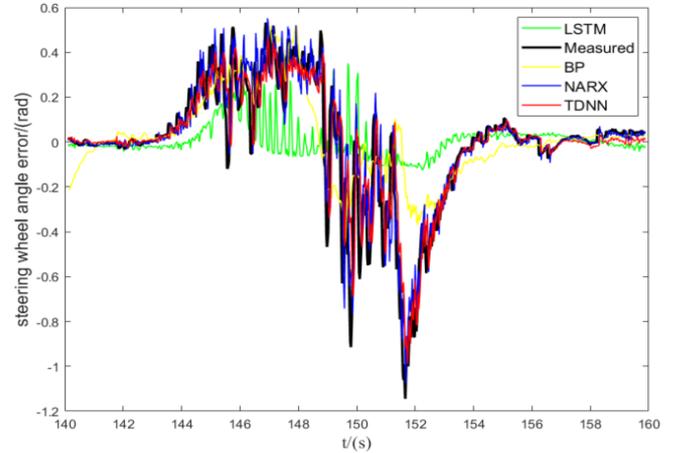

Figure 8.  Predicting comparision between different networks

As shown in the Fig. 8, each network was trained based on the same input variables: steering wheel angle, steering wheel torque, longitudinal velocity (with 5989 samples range). The other three networks have the same learning rate $l_R = 0.001$. The NARX and BP network both have two hidden layers with 8 nodes and 6 nodes. The LSTM network has one layer with 150 hidden units. We have trained different types of networks both for ten times, and every time the network starts from random parameters. In total, we get ten different sets of network parameters from the ten trainings, and we average the prediction results of them to reduce the randomness of the prediction performance. The average predicted results are shown in Fig. 9.

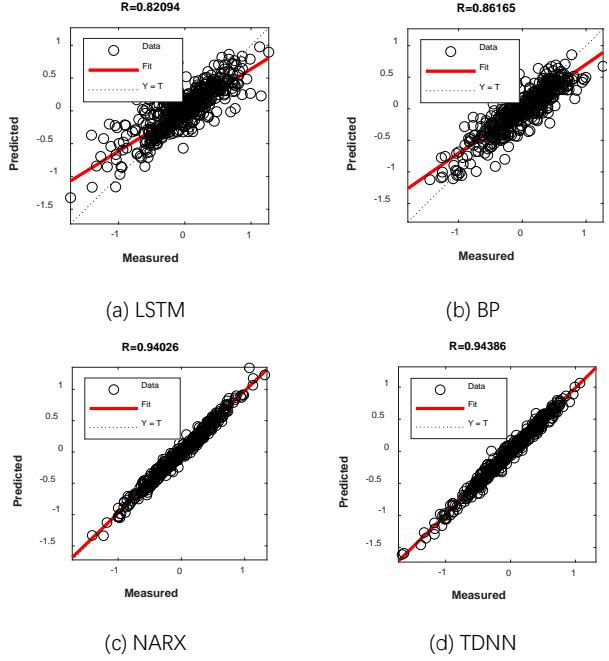

Figure 9. Scatter Plot of Predicted Obtained from Different Models

As we can see from Fig. 8 and 9, the results show that BP network has a weak ability for predicting nonlinear system. As the changing speed of the input signal gets larger, the network's prediction gets less accurate. The LSTM network will be easily overfitting because the complex structure of the network. Besides, the LSTM network has the longest training time, which is nearly three times longer than the other networks. Therefore, it is not an ideal network to implement in the compensator of the vehicle. NARX and TDNN network have similar predicting results. To further compare the performance of these two networks, we consider two coefficients to evaluate these two networks' predicting accuracy [12].

First, the correlation coefficient (CC) for evaluating accuracy is defined as:

$$CC = \frac{\sum_{t=1}^{N}[H_{get}(t) - \bar{H}_{get}][H_{pre}(t) - \bar{H}_{pre}]}{\sqrt{\sum_{t=1}^{N}[H_{pre}(t) - \bar{H}_{pre}]^2}\sqrt{\sum_{t=1}^{N}[H_{get}(t) - \bar{H}_{get}]^2}} \quad (6)$$

where $N$ is the number of the samples, $H_{get}(t)$ is the get by measuring steering wheel angle error in time t, and $H_{pre}(t)$ is the predicted steering wheel angle error in time t. The $\bar{H}_{get}$ and $\bar{H}_{pre}$ are the mean values of the measured and predicted data.

Second, the coefficient of efficiency (CE) for evaluating efficiency is defined as:

$$CE = 1 - \frac{\sum_{t=1}^{N}[H_{pre}(t) - H_{get}]^2}{\sum_{t=1}^{N}[H_{get}(t) - \bar{H}_{get}]^2} \quad (7)$$

We also extracted the straight and curved road scenario with different number of samples to increase the uncertainty of the training data. The modeling results with the evaluating indexes are shown in the following Table II.

TABLE II COMPARISION BETWEEN TDNN AND NARX

| Case | Samples number | Network | CC | CE |
|---|---|---|---|---|
| Straight lane | 5989 | TDNN | 0.895 | 0.874 |
| | | NARX | 0.924 | 0.901 |
| | 425 | TDNN | 0.791 | 0.836 |
| | | NARX | 0.889 | 0.921 |
| Curve lane | 5989 | TDNN | 0.981 | 0.961 |
| | | NARX | 0.921 | 0.912 |
| | 425 | TDNN | 0.872 | 0.851 |
| | | NARX | 0.893 | 0.901 |

According to data analysis in section III, the total error in straight-road cases is less than those in curved road. The main reason for the error may be the inaccuracy of the sensors or the disturbance from the environment. NARX network shows good modeling ability for the irregular data. As in curve road, the main contributing factor may be the time delay of the steering control system, the TDNN has good modeling ability for the apparent regular data. Furthermore, with the decrease of training data quantity, the predicting ability of TDNN network decreases. However, the NARX network's predicting ability remains almost the same.

Given the fact that the low-level control system has the problem of inaccuracy in U-turn control and the TDNN network is good at optimizing U-turn, the TDNN network, compared with NARX, is more suitable in our analysis. The existent control system (without compensated) usually has more steering wheel angle errors during U-turn than during straight route. To solve this problem, we can collect necessary data to ensure the training accuracy by modeling and compensating. In summary, TDNN is more stable and accurate and is adopted as the predicting network in our compensator.

*B. Model Validation*

To validate the prediction ability of the proposed data-driven method, we test our algorithm in simulation. We adopted CarSim as our simulation environment, which can be co-simulated with MATLAB SIMULINK.

In the simulation vehicle platform, the error between the steering wheel angle command input and the measured output is similar as our real vehicle platform. The error mainly includes two terms:

$$error = w_1 * error_{TD} + w_2 * error_{RD} \quad (8)$$

One is the $error_{TD}$ which is due to the characteristic of the time delay in the control system. The other is $error_{RD}$ which is due to the random disturbance of the environment and measurement inaccuracy of the sensors. The weighting values are $w_1 = 71.3\%$, $w_2 = 28.7\%$, which are calculated based on our system analysis results in Section III.

In the simulation scenario, the vehicle parameters are computed by the Multi-Rigid Body Dynamics and road function embedded in CarSim. We set the sampling period $T = 0.05s$, with the vehicle's main parameters the same as our testing platform Lincoln MKZ hybrid. According to the minimum turning radius of the testing platform, we set the front wheel angle $|\delta_f| < 0.4rad$, the initial course angle $\varphi(0) = 0$, the initial front wheel angle $\theta(0) = 0$, and the initial speed $(0) = 30km/h$. The double lane change scenario is commonly used to evaluate the steering control performance of a vehicle [11]. The scenario is with a total length of 200 meters and a double lane change is defined in Fig. 10.

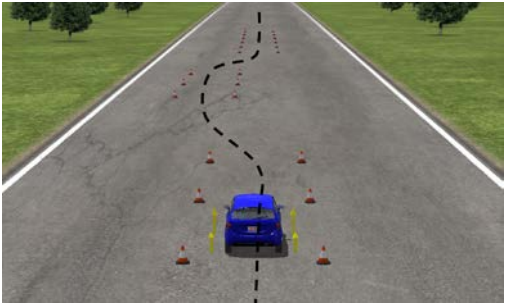

Figure 10. Simulation in Double Lane Change Scenario

The following Fig. 11 and Fig. 12 depicted the optimized results compared with the original results in 'double lane change scenario', the steering wheel angle performance becomes more and more smooth and stable, which demonstrates the improved performance of our controller.

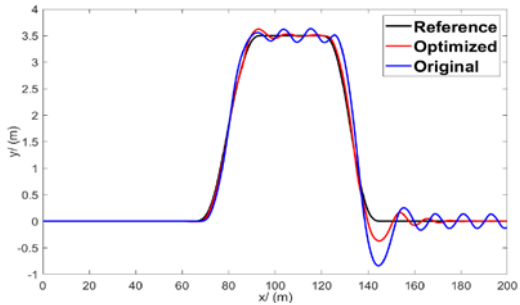

Figure 11. Comparision of Path Tracking Performance between the Optimized and Oringinal Method

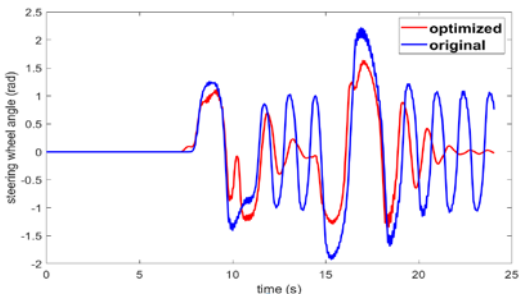

Figure 12. Comparision of Steering Wheel Angle Performance between the Optimized and Oringinal Method

## V. CONCLUSION

In this paper, we proposed a data-driven method for modeling and optimizing the controller of autonomous vehicles. From the analysis of the naturalistic driving data, we found out that 71.3% of the steering error was due to the time delay in the steering controller. We used TDNN network as our error compensation model by comparing the error predicting performances of the four typical networks (BP, NARX, LSTM, TDNN). Based on the error compensation model we developed feedforward control model which shows better path tracking performance. The maximum path tracking error after optimization was improved by 44.4% compared with the original one. The steering wheel angle oscillation was reduced 26.7% from the original data. In conclusion, our method shows the capability in improving the steering stability and control accuracy. Furthermore, our proposed approach also showed effectiveness in simulation scenario.

Our future work will look into the fluctuations caused by the uncertainties in real-world driving, and the controller's performance in low speed double lane changing situations.

ACKNOWLEDGMENT

The authors would like to thank Berkeley Deep Drive for funding support and Dr. Chen-Yu Chan and Prof. Chaoyang Jiang for their advice.

REFERENCES

[1] C. Badue, R. Guidolini, R. V. Carneiro, P. Azevedo, V. B. Cardoso, A. Forechi, L. F. R. Jesus, R. F. Berriel, T. M. Paixão, F. Mutz, T. Oliveira-Santos and A. F. D. Souza.(2019) Self-Driving Cars: A Survey. arXiv preprint arXiv:1901.04407, 2019.
[2] Silva, L. I., Magallan, G. A., De Angelo, C. H., and Garcia, G. O, "Vehicle dynamics using multi-bond graphs: Four wheel electric vehicle modeling," An Introduction to Signal Detection and Estimation. *New York: Springer-Verlag*, pp.2846-2851, 2008.
[3] Will, A., and ZëAk, S, "Modelling and control of an automated vehicle," *Vehicle System Dynamics*, vol. 27, no. 3, pp.131-155, 1997.
[4] Smith, D. E., Starkey, J. M., and Benton, R. E, "Nonlinear-gain-optimized controller development and evaluation for automated emergency vehicle steering," *In Proceedings of the 2000 IEEE American Control Conference*, pp.3586-3591, 2000.
[5] P. Zhao, J. Chen, Y. Song, X. Tao, T. Xu, T. Mei, "Design of a control system for an autonomous vehicle based on adaptive-PID," *International Journal of Advanced Robotic Systems*, pp.44, 2012.
[6] Z. Li, Y. Li, C. Yang and N. Ding, "Motion control of an autonomous vehicle based on wheeled inverted pendulum using neural-adaptive implicit control," in *IEEE/RSJ International Conference on Intelligent Robots and Systems on. IEEE*, 2010, vol.25, pp.133-138.
[7] Alexa, O., Ilie, C. O., Viläƒ U, R., Marinescu, M., and Truta, M, "Using neural networks to modeling vehicle dynamics," *Applied Mechanics and Materials*, pp.133-138, 2014.
[8] P. Wang, C. Chan, "A Reinforcement Learning Based Approach for Automated Lane Change Maneuvers," in *2018 IEEE 29th Intelligent Vehicles Symposium (IV)*, pp.1379-1384.
[9] Jolliffe, I. T, "Principal Component Analysis," *second edition Springer-Verlag*, 2002, pp.10-20.
[10] Quigley, M, "ROS: an open-source Robot Operating System," In Proceddings of 2002 *IEEE ICRA Workshop on Open Source Robotics vol.3, no 3.2, pp.5*, 2009.
[11] Double lane-change ISO:https://www.iso.org/obp/ui/#iso:std:67973:en
[12] Yen-Ming Chiang, Ruonan Hao, Nannan Li, and Jian Wang, "Water Level Forecasting by Static and Dynamic Neural Networks," *Journal of Tianjin University (Science and Technology)*, vol.3, pp. 25-34, 2017
[13] L. Xin, P. Wang, C. Chan, J. Chen, S. E. Li and B. Cheng, "Intention-aware Long Horizon Prediction of Surrounding Vehicle Trajectory using Dual LSTM Networks," *The 21st IEEE International Conference on Intelligent Transportation Systems (ITSC)*, 2018: 1441-1446
[14] C. Xi, T. Shi and J. Chen, (2019). "A data driven approach for motion planning of autonomous driving under complex scenario," arXiv preprint arXiv:190408784.
[15] P. Wang, C. Chan, H. Li. "Automated Driving Maneuvers under Interactive Environment based on Deep Reinforcement Learning," *Transportation Research Board (TRB)*, Washington D.C., US, 2019